\documentclass[10pt,twocolumn,letterpaper]{article}

{\noindent\ignorespaces\color{#1}\textbf{#2:}}%
{\par\noindent\ignorespacesafterend}

\newcommand{\eg}{\emph{e.g.,}}
\newcommand{\ie}{\emph{i.e.,}}

\usepackage{wacv}
\usepackage{times}
\usepackage{epsfig}
\usepackage{graphicx}
\usepackage{amsmath}
\usepackage{amssymb}

\usepackage{gensymb}
\usepackage{balance}
\usepackage{algorithm}
\usepackage[noend]{algpseudocode}
\usepackage{placeins}

\usepackage{siunitx}
\usepackage{tabu}
\usepackage{tabularx}
\usepackage{tablefootnote}
\usepackage{threeparttable}
\usepackage{mathtools}
\usepackage{epstopdf}
\usepackage{float}
\usepackage[T1]{fontenc}
\usepackage{enumitem}
\usepackage[percent]{overpic}
\usepackage{color}

\wacvfinalcopy

\graphicspath{ {plots/} {diagrams/} {figures/}}

\def\wacvPaperID{104} 

\ifwacvfinal\fi
\begin{document}

\title{Autonomous Curiosity for Real-time Training Onboard Robotic Agents}



\author{Ervin Teng \\
Carnegie Mellon University\\
{\tt\small ervin.teng@sv.cmu.edu}
\and
Bob Iannucci \\
Carnegie Mellon University\\
{\tt\small bob@sv.cmu.edu}
}


\maketitle
\ifwacvfinal\thispagestyle{empty}\fi

\begin{abstract}

Learning requires both study and curiosity. A good learner is not only good at extracting information from the data given to it, but also skilled at finding the right new information to learn from. This is especially true when a human operator is required to provide the ground truth---such a source should only be queried sparingly. In this work, we address the problem of curiosity as it relates to online, real-time, human-in-the-loop training of an object detection algorithm onboard a robotic platform, one where motion produces new views of the subject. We propose a deep reinforcement learning approach that decides when to ask the human user for ground truth, and when to move. Through a series of experiments, we demonstrate that our agent learns a movement and request policy that is at least 3x more effective at using human user interactions to train an object detector than untrained approaches, and is generalizable to a variety of subjects and environments.

\end{abstract}




\section{Introduction}
\label{sec:intro}

Effective learning is a complex set of inter-related processes involving both \textit{study} and \textit{curiosity}. In order to learn about a particular subject, we as students seek to extract as much knowledge as possible from the information at hand.  We also actively seek new information that would complement what we already have---providing new insights and clarification of old ones---helping us to formulate a clear understanding of the subject.  In many cases, we have help from a teacher or mentor.  This help is often structured (\eg{} in time-limited classroom settings, mentoring sessions) and finite.  A good student will use the teacher's time judiciously--learning what he or she can independently, and then asking for help when necessary. A great student studies effectively, exhibits strong curiosity, and learns how to take the best advantage of teachers and mentors.

In this work, we apply these observations to the training of an object detection algorithm embedded in a robotic agent.  We take as our example student a small unmanned aerial system (a ``drone'') that carries onboard such an object detector that must be trained online.  While many have explored pre-trained object detectors embedded in various robotic agents, we go further to consider the need for incremental online training and how the robotic agent can exhibit (a) curiosity and (b) the ability to selectively ask for assistance from a human trainer in ways that maximize the benefit of both such actions.

Consider this scenario: our drone and its object detector are on a mission that focuses attention on a particular subject individual.  The individual now steps into a car.  A human operator would naturally turn attention to tracking the car.  But our drone would need to have its focus re-directed and, in fact, it may need to be re-trained to detect that particular car.  The real-time nature of the mission precludes a stop-and-retrain approach (\eg{} gathering new images of the subject car, building a new training set, re-training the object detector, and re-deploying the detector to the drone). Assuming our detector is a convolutional neural network, we can imagine some means for in-flight re-training with input from a human operator.

\begin{figure}[t] \centering
    \includegraphics[width=\linewidth]{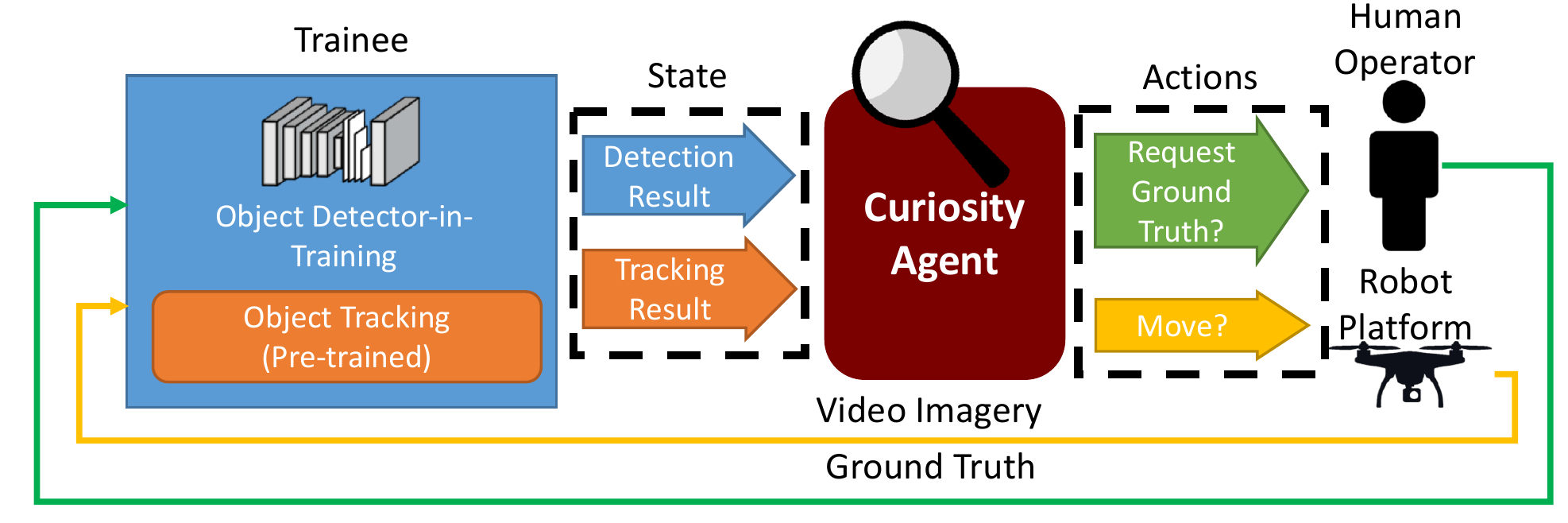}
    \caption{Diagram of how a curiosity agent can assist real-time online training of an object detector onboard a robotic platform (\eg{} a drone). The curiosity agent receives feedback from the trainee, which consists of the learning object detector and an object tracker which produces training events as long as the subject is tracked. It is then responsible for giving the trainee new information to track and subsequently train on, whether by asking a human operator for ground truth, or by moving the robot and camera to a different viewing position.}
    \label{fig:clickbait_overview}
\end{figure}

But have we just moved an extensive offline training problem to the shoulders of the trainer?  Extensive input from the trainer---akin to the density of training examples in an offline training set---might simply be impractical in a real-time setting.  We postulate that it is possible to amplify the \emph{benefit} of real-time human input in such a situation by improving how a given stream of images is \textbf{studied} (processed) and in how the robotic agent can alter the stream of images by \textbf{exhibiting curiosity}.  This amplification, if successful, could substantially reduce the cognitive load imposed on the trainer over traditional approaches.


Improving the benefit of given training events (\textbf{study}) has been widely explored~\cite{Teng2018ClickBAIT:Networks,Papadopoulos2017,Papadopoulos2016WeVerification}.
Active learning partially addresses the question of \textbf{curiosity} by selecting (subsampling) which datapoints would be most advantageous to annotate~\cite{Settles2010, Wang2017,Fang2017}.  We extend such thinking by coupling the learning process with movement---giving the learner the ability to move its sensors (\eg{} camera) to seek out particularly beneficial training data.

In this paper, we advance a new concept that we call \textbf{autonomous curiosity}: a method for coupling incremental training to the navigation of an autonomous system in such a way that the effectiveness of human training assistance is amplified. By combining autonomous curiosity with state-of-the-art means to extract information from sensors, we create a new approach for real-time incremental training in which the learner exercises a kind of ``good judgement'' in deciding when to seek input from a human trainer.



We claim the following contributions in this work:
\begin{itemize}[noitemsep]
\item An approach for measuring and systematically improving the the \emph{training benefit} of human-trainer interactions through guided camera movement, employing deep reinforcement learning and procedurally-generated 3D simulations.
\item Validation of this approach through both simulation and flight-based field studies. We demonstrate an increase in mean incremental training benefit of more than 3x as compared to an untrained baseline.
\end{itemize}

Section~\ref{sec:background} gives an overview of the problem domain and related work. Section~\ref{sec:problem} formulates the training of an object detector onboard a drone platform as a reinforcement learning problem. Section~\ref{sec:architecture} describes the architecture of the curiosity agent itself and its reward structure. Section~\ref{sec:experiments_simulation} describes how the agent is trained and evaluated, and Section~\ref{sec:results_simulation} and \ref{sec:experiments_field} show the results of evaluating the curiosity agent on simulated scenes and real-world imagery, respectively. Section~\ref{sec:extensions} explains how this work could be extended to multi-class detection and more complex motions.

\section{Background and Prior Work}
\label{sec:background}

\subsection{Problem Domain}

In order to allow for mid-mission re-training of deep learning models, we must enable real-time human-in-the-loop annotation and training. The annotations generated by the user, in conjunction with the video frame they correspond to, become the training set for the object detector. If we assume the detector itself has a static architecture~\cite{Liu2016,Redmon2015} with a fixed sample efficiency, the remaining figure-of-merit for our system is the model improvement garnered each time the user has to interact with and provide ground truth to the system.

In this work, we will examine the subproblem of real-time training for single-class single-instance detection. We define a training \textit{episode} as a single training session with one subject object. If we divide the episode into discrete time-steps, at each time-step $i$, the user may or may not provide ground truth to the system. This forms a vector $U$ of user interactions:
\begin{equation}
U = (u_1, u_2, ..., u_n),   u_i = \begin{cases}
      1 & \text{user int. on frame $v_i$} \\
      0 & \text{otherwise}
   \end{cases}
\end{equation}
The metric in question, then, is the average \textit{incremental training benefit}~\cite{Teng2018ClickBAIT:Networks} of each user interaction during an episode. Qualitatively, incremental training benefit is a measure of the increase in model performance (average precision in this case) that can be attributed to a particular user interaction. For a fixed number of time-steps within an episode, we can compute the average ITB from time-step $i$ to $j$ with:

\begin{equation}
\overline{ITB}_{u_{(i,j)}} = \frac{P_j - P_i}{\sum_{x=i}^{j} u_x}
\label{eqn:meanitb}
\end{equation}
where $P_i$ is the performance (\ie{} the AP) of the trainee model at time-step $i$, and $P_0$ is reserved for the performance of the object detector before any online training.

\subsection{Related Work}

\textbf{Active perception}, \ie{} guided exploration of visual data, has been applied to object localization in single images~\cite{Ranzato2014,Mathe2016,Caicedo2015}, activity recognition in videos~\cite{Yeung2016End-to-endVideos}, and object recognition for physical 3D objects through robotic manipulation~\cite{Jayaraman2016Look-aheadMotion,Malmir2015}. In these applications, the active agent itself performs the localization, detection, or classification task at hand, and results in being application or class-specific. In~\cite{Jayaraman2018}, the authors present an active agent that \textit{learns to learn}, \ie{} that learns to teach itself about an object or scene. Our approach differs from this concept in two distinct ways: first, that it is easily separable, in that the active curiosity agent trains a separate model for future use \textit{independent of} the agent, and second, in that it learns from a human-in-the-loop rather than solely from its own observations.

\textbf{Robotic exploration} and navigation, whether through explicit~\cite{Martinez-Cantin2007ActiveUncertainty} or learned~\cite{Mirowski2017LearningEnvironments,Pathak2017Curiosity-drivenPrediction,Tai2016MobileLearning,Guisti2016} means, share the same concept of moving a robotic agent to explore a space or scene. However, the goal (and likewise, the reward), is different. In our approach, we want to explore not exhaustively but only that which is \textit{useful for training a learner}.

\textbf{Active learning}, unlike active perception, starts with the premise that there is a human oracle/annotator who is only willing to annotate a finite amount of data. Traditional active learning attempts to maximize the performance of the model trained given a fixed number of annotated datapoints, and a number of heuristics, including expected model change, expected loss reduction, or core-set coverage, are used to determine the value of each annotation request~\cite{Settles2010,Yao2012InteractiveDetection,Sener2017}. Reinforcement learning agents~\cite{Fang2017} can be used for active learning in streaming applications where the whole dataset is not known \emph{a priori}.

Other than data selection, weakening annotation supervision~\cite{Papadopoulos2017,Papadopoulos2016WeVerification,Jain2016ClickClicks,Mettes2016SpotProposals,Bearman2016WhatsSupervision,Xu2016,Teng2018ClickBAIT:Networks} is a common method to reducing human annotation time. This body of work is complimentary to our approach, as we do not address \textit{how} the human provides the annotation, just \textit{what} is annotated.

Our approach, autonomous curiosity, combines active learning and active perception in a single agent to improve the effectiveness of user interactions through both data selection and self-driven exploration.

\section{Problem Setup}
\label{sec:problem}
In order to tackle the problem of improving training benefit using autonomous curiosity, we map the general problem of real-time training in a 3D environment onto a simplified scenario with constrained motion. Note that we are not autonomously following the subject. While ample literature~\cite{Luo2018End-to-endLearning,Bayoumi2015EfficientLearning,Pestana2014ComputerVehicles} exists on how to enable robots to autonomously follow chosen subjects, we focus on enabling \textit{autonomous learning} through motion.

In this formulation, there are two learned components. First, an object detector onboard a drone (the \textit{trainee}) that must be fine-tuned to detect a particular object (the subject) in the environment. Second, the \textit{curiosity agent} is responsible for moving the drone around and requesting ground truth input from a human operator to train the trainee. In this work, we focus on the curiosity agent, and treat the trainee as part of the problem setup.


\subsection{Environment}

\begin{figure}[t] \centering
    \includegraphics[width=0.9\linewidth]{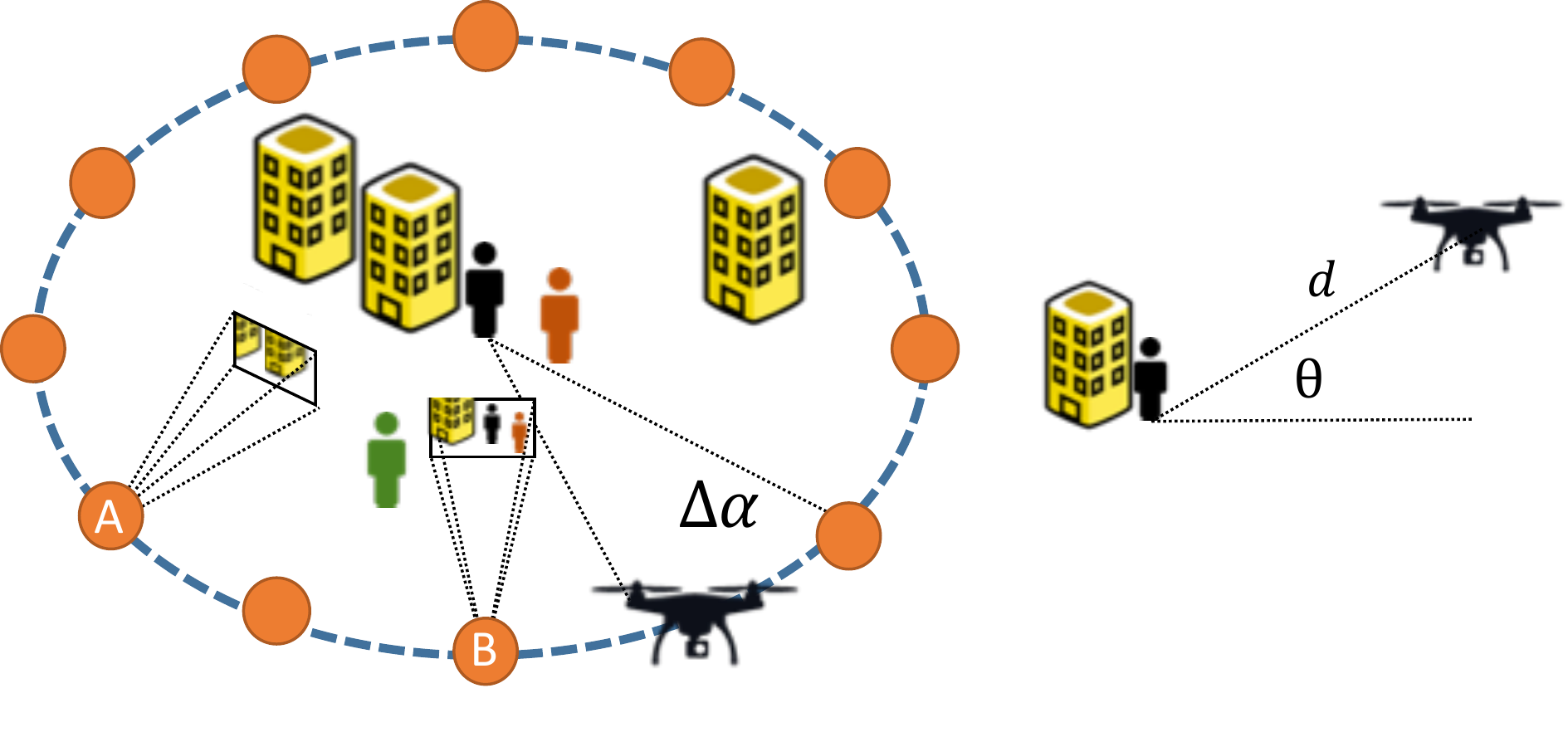}
    \caption{(Left) Diagram of orbit around subject (person in the center). The drone can exist at any one of discrete points along the orbit, and movement left or right progresses to the next discrete point. These points are separated by an angle $\Delta \alpha$. The drone's camera is always pointed towards the subject in the center. At each point, the drone has some view of the subject, where it may be obstructed (position A) or contain confusing false positives (position B). Each view from a position, therefore, has different impact on training. (Right) Side view of drone relative to subject.}
    \label{fig:orbit}
\end{figure}

We constrain the movement of the drone to an orbit around the subject, as shown in Figure~\ref{fig:orbit}. An orbit is chosen as the first application of autonomous curiosity as it is has a sufficiently simple range of motion (just left and right), but also has the property that different positions on the orbit can have vastly different views of the subject. The orbit itself is positioned relative to the subject with respect to two parameters, the elevation angle $\theta$ and the line-of-sight distance $d$ from the subject to the drone. In our orbits, we set $\theta = 30\degree$, and $d$ is variable depending on the subject.

We also discretize the orbit path to constrain the otherwise infinite set of possible training images that can be gathered on the orbit. Each point along the orbit path is separated by a fixed angle $\Delta \alpha$. We chose this angle to be $\Delta \alpha = 12 \degree$---roughly 1/3 of the $37.2\degree$ FOV of a GoPro Hero 4 Black in Narrow mode, cropped to a square (the imaging setup~\cite{Teng2018ClickBAIT:Networks} used in our field experiments).

We will subsequently refer to a generated scenario with a fixed subject and some obstructions as a \textit{scene}, an image of the subject from a point on the orbit as a \textit{view}, and the set of all possible views of the subject (one for each point in the orbit) as the \textit{exploration space}. 

\subsection{Trainee}

The \textit{trainee} is an online training system for object detection with the same architecture, data augmentation, and object tracking strategy as the one presented in~\cite{Teng2018ClickBAIT:Networks}. The object detector (SSD) is trained from ground truth bounding box data given by the human user, and object tracking is used to interpolate additional ground truth information between user inputs.

\label{sec:trainingsys}
\subsubsection{Object Detector}
\label{sec:ssd}
We chose the single shot object detector (SSD)~\cite{Liu2016} for its high frame rate as compared to region proposal methods. We follow closely the configuration used by~\cite{Teng2018ClickBAIT:Networks}, a VGG16-SSD model with 300x300 pixel input, re-structured as a two-class (positive and negative) model from the standard PASCAL VOC 20 classes.

Each episode (\ie{} each series of training events and movement actions within a single exploration space) begins with the model above. During the episode, annotated views are used to initiate incremental training rounds, \ie{} gradient updates, on the view and augmented versions of the view, and the model learns subject-specific traits.



\subsubsection{Object Tracking}
\label{sec:optical_flow}
To further reduce required user interactions, \cite{Teng2018ClickBAIT:Networks} used an object tracker to generate additional training rounds, without user interaction, as long as the subject remains tracked. We adopted the same strategy, using a state-of-the-art real-time tracker based on correlation filters and a Siamese network~\cite{Valmadre2017}. To stop a runaway tracker causing erroneous training, during the search phase, if no image patch matches with a correlation that exceeds a threshold $t_{tr}$, a tracking failure is reported and tracking is terminated. $t_{tr} = 5\text{e-}5$ was empirically determined to perform well for this application. Note that even with a conservative threshold, object trackers can drift---it is the responsibility of the agent to request a new ground truth bounding box when the tracking result is no longer useful.

\subsection{Actions}

The curiosity agent must both navigate around the orbit as well as request from the user ground truth for the trainee. Actions taken by the curiosity agent should influence the drone's position in the orbit, the state of the trainee, or both. The curiosity agent can take advantage of the trainee's object tracking abilities to avoid asking for ground truth at every point along the orbit.

At each time-step $i$, the agent can take one of four actions from the set of actions $A=\{0,1,2,3\}$:
\begin{itemize}[noitemsep]
\item Don't move. If object tracking is initialized, perform a training round.
\item Request user interaction, which produces a ground truth bounding box. Ground truth is then used for a training round and to initialize object tracking.
\item Move left. The drone moves left one spot in the orbit. If object tracking was successful prior to the move, we train on the previous view during the transit.
\item Move right. The drone moves right one spot in the orbit. If object tracking was successful prior to the move, we train on the previous view during the transit.
\end{itemize}Algorithm~\ref{alg:curiosityagent} shows the process during an episode. The curiosity agent can interact with the user through requestGroundTruth, where the user is sent the current drone view of the subject and receives a bounding box ground truth from the user. In addition, at each time-step where ground truth is not received, the current tracked bounding box, if present, is used to initiate a training round for the trainee.

\begin{algorithm}
\caption{Curiosity agent during an episode. At each time-step, the agent can either not move, move the drone, or request ground truth. In addition, if an object is being tracked by the trainee, a training event is initiated for the trainee. If updateTracker fails, $b_{track}$ is set to None.}\label{alg:curiosityagent}
\begin{algorithmic}[1]
\State $\alpha \gets 0$ \Comment{Position in orbit}
\State $b_{track} \gets $None \Comment{Tracking BBox}
\State $v \gets $getCurrentView($p_{orbit}$)
\For{$i$ = (0,$n$)}
  \State $a_i \gets $chooseAction(A) \Comment{Agent chooses next action}
    \If{$a_i = 0$} \Comment{Don't move}
    	\If{$b_{track}$ is not None}
        \State trainingRound($v$, $b_{track}$)
        \EndIf
 	\ElsIf{$a_i = 1$} \Comment{Request User action}
    	\State $b_{track} \gets $requestGroundTruth($v$)
        \State trainingRound($v$, $b_{track}$)
    \ElsIf{$a_i = 2$} \Comment{Move left}
    	\If{$b_{track}$ is not None}
        \State trainingRound($v$, $b_{track}$)
        \EndIf
        $\alpha \gets \alpha + \Delta \alpha$ \Comment{Change orbit position}
    \ElsIf{$a_i = 3$} \Comment{Move right}
    	\If{$b_{track}$ is not None}
        \State trainingRound($v$, $b_{track}$)
        \EndIf
        $\alpha \gets \alpha - \Delta \alpha$ \Comment{Change orbit position}
    \EndIf
    \State $v \gets $getCurrentView($p_{orbit} $)
    \State $b_{track} \gets $updateTracker($v_{i+1}$)
\EndFor
\State \textbf{return}
\end{algorithmic}
\end{algorithm}

Note that in this formulation of the problem, we make the strong assumption that each of these actions has roughly the same \textit{time} cost. Our drone platform is a 3DR Solo with an NVIDIA Jetson TX2~\cite{Teng2018ClickBAIT:Networks}---developed to be GPU-accelerated, compact, and low cost (<\$1000). For this platform moving 5m/s with a line-of-sight distance $d$ of 60m, both a training round and movement to the next discrete point in the orbit take approximately 2.5s (inference is 4-5 fps). Extension to variable time-costs per action is left for future study.

\section{Approach}
\label{sec:architecture}

\label{sec:activelearning}

Rather than designing explicit rules for navigation, we elected to realize the curiosity agent with deep reinforcement learning and offline training, hypothesizing that training across a range of synthetic scenes would yield a generalizable curiosity policy. This section describes the design of the reward function, the representation of state for the agent, and the architecture of the agent itself.

\subsection{Reward Function}

At a high level, the goal of the curiosity agent is to train the best model with the least amount of user input. More specifically, it must execute a sequence of actions (Section~\ref{sec:problem}) that, for a fixed time budget $T$, produces the best trainee using the least number of ground truth requests. We must capture two elements in the reward function: the positive reward of improving the trainee's detection ability, and the negative reward of having to request ground truth from the user. We linearly combine both these elements as a single weighted reward function:
\begin{equation}
\mathcal{R}_{E} = w_p\mathcal{R}_{l} + (1-w_p)\mathcal{R}_{gt}
\label{eqn:reward}
\end{equation}
where $w_p$ is the weight of the positive learning reward, and can be regarded as an application-specific design parameter.

\textbf{Positive Learning Reward:} During training, we would like to reward the agent for producing a performant trainee. For an object detector, average precision (AP) can be used as a metric---but over what test set? Given the full exploration space around the orbit, the best learner performs well on \textit{all of the views}, with the hope that this will generalize beyond the scene. The curiosity agent must find some subset of these positions, and order in which to visit them, that would train the trainee without exhaustive search. This is similar to the concept of core-set~\cite{Sener2017} in active learning, \ie{} the smallest set of datapoints which can be used to train a learner to perform well on the entire set.

Furthermore, to increase the density of rewards, we want to reward the agent after each training event. Rather than using absolute AP over the exploration space, we use the incremental gain in AP for that time-step, \ie{} the \textit{incremental training benefit}~\cite{Teng2018ClickBAIT:Networks} of any training event that happened during that time-step. The reward function after time-step $i$, $0\leq i < T$ is given by the capped difference in training objective loss
\begin{equation}
\mathcal{R}_{l} = \min\{\max \{10 (P_i - P_{i-1}),-1\},1\}
\end{equation}
Where $P_i$ is the performance (AP) of the object detector trainee, trained up to time-step $i$, over the entire exploration space. We multiply the difference by a factor of 10 to bring the typical difference in AP, given the hyperparameters of our trainee, closer to 1.

\textbf{Negative Asking Reward:} If we purely rewarded the agent on improving trainee performance, it would quickly learn that asking the user to interact and provide ground truth every time-step would produce the cleanest training data. This goes against our goal of increasing the effectiveness of each user interaction. To encourage the curiosity agent to request user ground truth only when needed, we assign a \textit{negative} reward every time a ground truth request is made. Therefore, when $a_i = 1$,
\begin{equation}
\mathcal{R}_{gt} = -1
\end{equation}

\subsection{State Representation}

What information (\ie{} state) would the curiosity agent require to maximize the reward function? One approach would be to use the current view as state. An effective teacher, however, should not blindly go through training material---it needs some feedback from the trainee. For instance, training on a view for which the trainee is already performing well would not have much benefit. In addition, as the object tracker is not perfect, we would like the curiosity agent to know what is currently being used as ``ground truth'' for training.

\begin{figure} \centering
    \includegraphics[width=0.8\linewidth]{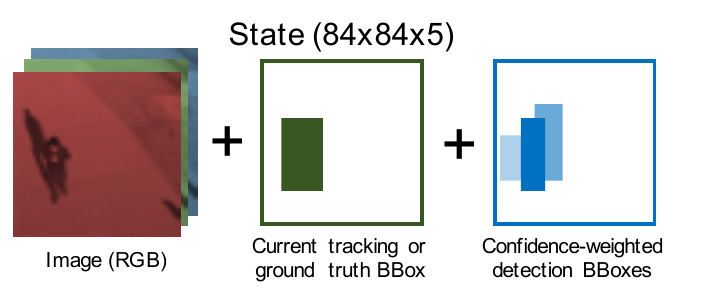}
    \caption{Matrix state representation for Q-Network. The RGB channels of the view are concatenated with representations of the current tracked bounding box and the current output of the trainee object detector, respectively. The output of the object detector consists of filled boxes whose magnitude is dependent on the confidence of the output.}
    \label{fig:staterep}
\end{figure}

\textbf{Trainee Feedback: }In training an object detector on a real-time video feed, a human would observe the presence and confidence of bounding boxes to monitor the training process. To incorporate both the behavior of the trainee and the output of the object tracker in the state representation, we concatenate two additional channels to the image view (shown in Figure~\ref{fig:staterep}), representing the state as an 84x84x5 matrix containing:
\begin{itemize}[noitemsep]
\item The current view $v$, RGB channels.
\item Filled-in bounding box (value $=255$) of the current ground truth BBox (as provided by the user) or tracked BBox, if object tracking is active.
\item Confidence-weighted BBox (value $=255*c$) of the detected bounding boxes, drawn in order of lowest confidence first (higher confidence boxes overlap and cover lower ones). Threshold for SSD output is 0.3.
\end{itemize}

\textbf{Positional Information: }One additional piece of information is useful for the curiosity agent: where it is on the orbit, so that it knows when it has revisited a position. We encode this information as a scalar between 0-1, with $p_i = \alpha / 360$, where $\alpha$ is the heading angle from the subject to the drone on the orbit.


\subsection{Curiosity Agent Architecture}

\begin{figure}[!htb] \centering
    \includegraphics[width=0.8\linewidth]{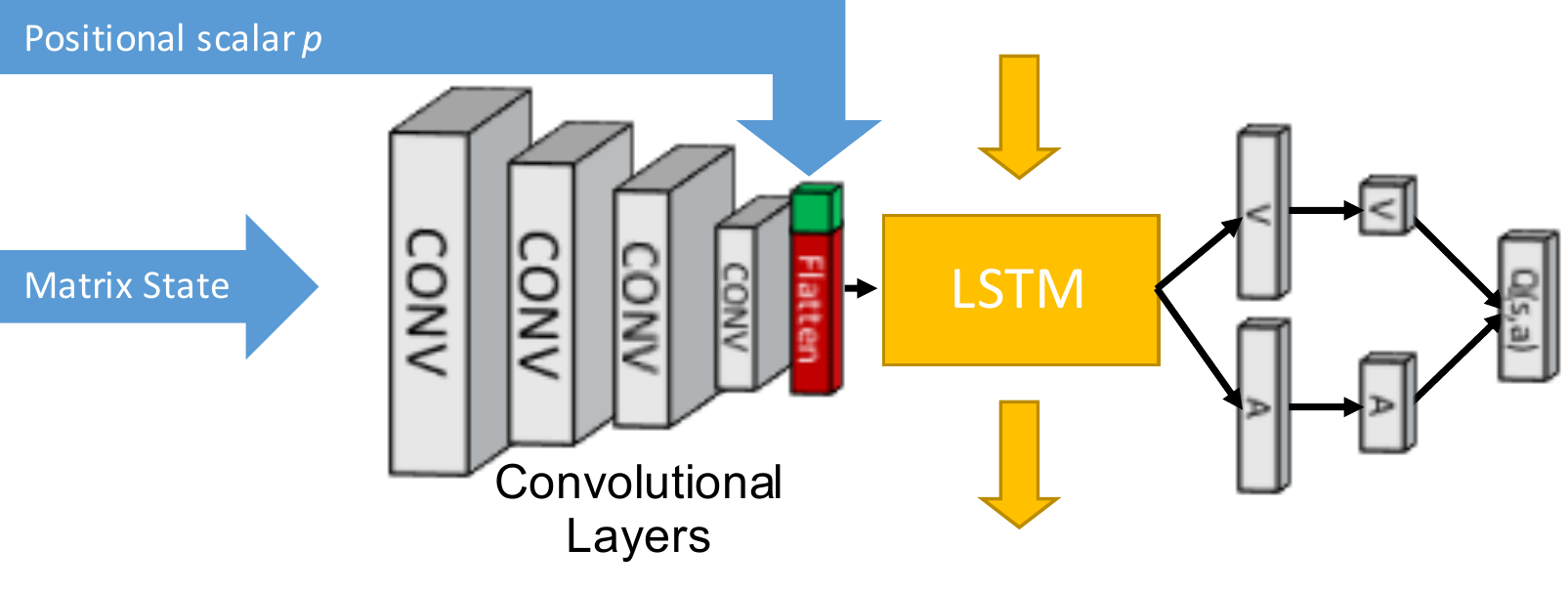}
    \caption{Structure of deep Q network with concatenated state. An LSTM gives temporal notion to the network, and a dueling setup adds additional stability during training. The feature extractor is composed of 4 convolutional layers having respectively, 32, 64, 64, and 512 output channels with filters of sizes 8x8, 4x4, 3x3 and 7x7 and strides 4, 2, 1, and 1.}
    \label{fig:DQN}
\end{figure}

We now have a discrete set of four actions and a state that consists of two parts, an 84x84x5 matrix as well as a scalar encoding the drone's position. Because our action space is discrete, we chose a value-based approach for its sample efficiency. As no one view contains enough information to observe the entire exploration space, our problem is \textit{partially observable}. Therefore, we use a deep recurrent Q-network architecture as presented by~\cite{Hausknecht2015DeepMDPs}, which has shown success in partially observable 3D environments~\cite{Lample2016PlayingLearning}. To incorporate the scalar state, the convolutional output is concatenated with the $p_i$ value as shown in Figure~\ref{fig:DQN}. A dueling approach~\cite{Wang2016DuelingLearning} and double-Q learning~\cite{vanHasselt2016DeepQ-learning} are applied to improve stability during training and allow for faster learning rates.

\section{Training and Experiments}
\label{sec:experiments_simulation}
To both train and validate our curiosity agent without overfitting, we required a large number---more than 14,000---of unique orbit scenes. The variety of scenes would directly impact the generalization of the learned policy. As this would be prohibitively difficult to do in the field, we use procedurally generated simulation to construct the scenes. Here, we describe the simulation environment, the training process, and several baseline training strategies we compared against our agent.

\subsection{Simulation Environment}
As we want scenes to vary in both appearance and geometry, we procedurally generate them using the Unity 3D game engine, which interacts with the Tensorflow DQN network through the ML-Agents API~\cite{Juliani2018Unity:Agents}. At the beginning of each episode, we create a scene by first choosing one of six different subjects. A number of obstructions $n_{obs}$, from eight different types of varying height, are then placed on the ground in a region around the subject within the diameter of the orbit. This is done so that none of the obstructions end up overlapping either the camera or the subject. For a standoff distance $d = 60$ (simulated) meters, we set the inner radius of the placement region to 10m and the outer to 35m. We also perturb the center point of the orbit by up to 5m (conservative estimate of GPS accuracy) to simulate inaccuracy in aiming the camera.

In~\cite{Sadeghi2016CAD2RL:Image}, the authors underscore the importance of randomization during simulation---and that a depth perception model trained with random, unrealistic colors and textures will outperform one that has been trained purely on realistic scenes by about 10\%. In our simulations, we randomize both the color of the obstructions and the subject to encourage generalizability in the trained agent.

\begin{figure}[!htb] \centering
    \includegraphics[width=0.9\linewidth]{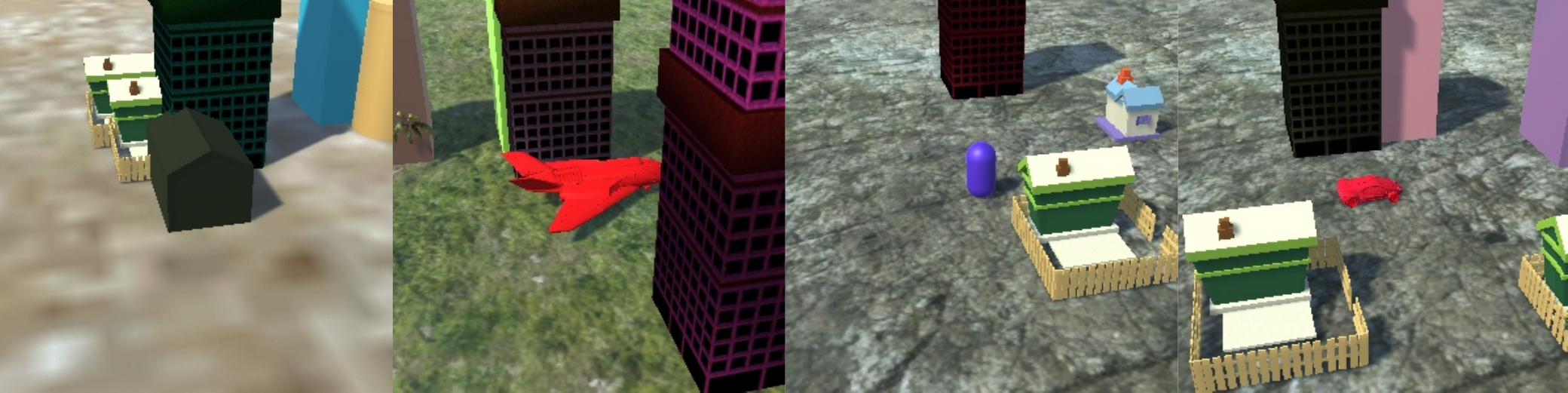}
    \caption{Views of generated scenes for 4 of the 6 types of subjects (house, plane, cube, capsule, car; Not shown: cube, sphere).}
    \label{fig:sim_screenshots}
\end{figure}

As described in Section~\ref{sec:problem}, we divide the orbit into discrete views, and allow the drone to navigate around the orbit. The virtual camera, which mimics our real-world hardware setup, is set to have a $37.2\degree$ FOV. Figure~\ref{fig:sim_screenshots} shows examples of the environments that can be generated, with the sample subjects in the center of each view.

\subsection{Training in Simulation}

As is standard for DQN networks, experience replay was used to update the DQN, of which only the latter half of replay states are updated~\cite{Lample2016PlayingLearning}. For each training episode, we give the curiosity agent a fixed time budget of $T=100$, which balances the the time available to explore the space to the time an agent could waste training a fully-trained ($P_i = 1.0$) trainee. Typically, a trained agent can reach trainee performance $P_T > 0.8$ after 100 time-steps. To stabilize training, we use curriculum learning~\cite{Bengio2009CurriculumLearning}, gradually increasing $n_{obs}$ from 10 to 20, where 20 produces scenes which are mostly obstructed. During training, we chose $w_p$ to be 0.8, the smallest value (fewest user interactions) without a significant degradation in trainee performance.

\subsection{Baseline Strategies}
\label{sec:baseline}

We will judge our curiosity agent against the following 4 baselines. These strategies all have the requirement that they must act, at most, on the same state information and with the same set of actions as the curiosity agent.

\textbf{Random:} The agent chooses a random action at each time-step. This strategy does not consider state information.

\textbf{Orbit with Tracker-only:} If our goal is to achieve the best trainee performance over the entire exploration space, logically, we would want to train on each and every view one at a time. This can be achieved by constantly moving in the same direction, forming a full orbit of the subject. As in~\cite{Teng2018ClickBAIT:Networks}, we can avoid asking for user input at each time-step by using object tracking. When a user annotates a frame with a ground truth bounding box, the object tracker is initiated. The agent then always chooses $a_i = 2$ (move right and initiate training round if the tracker is successful). When a tracking failure is detected, ground truth is again requested from the user. This ensures that subsequent similar views are trained on without user interaction.

\textbf{Jump $n$:} A one-step-at-a-time orbit has the problem of requiring user input at every time-step when the subject is obstructed. Furthermore, there is no mechanism for correcting a ``runaway'' tracker that produces incorrect training events. We can mitigate these problems by observing that views far away from each other are more likely to produce new information. In this strategy, we always move $n$ steps to the right before asking for ground truth again. For comparison, we select $n = 7$ and $n = 10$, with 7 being a non-integer fraction of 30 (the number of views in the exploration space), so that multiple orbits cover different views.

\textbf{Histogram Match $t_{hist}$: } The \textbf{jump} strategy intentionally uses \textit{distance} as a determiner of view uniqueness. In this strategy, we use the similarity of the histogram of the two images to determine when to ask the user for input. Rather than asking the user for ground truth after a fixed number of movement steps, after each movement step, we compute the histogram of the view. We then compute the correlation of the histogram with the one of the view last annotated, and ask for a user interaction if it drops below a threshold $t_{hist}$, which is scene-specific.


\section{Results - Simulated}
\label{sec:results_simulation}

We evaluated our curiosity agent both in respect to average incremental training benefit (Equation~\ref{eqn:meanitb}) over the entire episode, as well as the temporal behavior of the trainee's performance during each time-step. We evaluated each strategy across 15 randomly generated environments containing between 10 to 15 obstacles. By fixing the seed at the beginning of each evaluation session, every strategy receives the same 15 environments, even though each environment is unique. For each evaluation episode, we give the agent a time budget of $T=100$, as in training. For realism, actions which do not take any time, \ie{} not moving and not training, do not count towards the time budget.

\begin{figure}[!htb] \centering
    \includegraphics[width=1\linewidth]{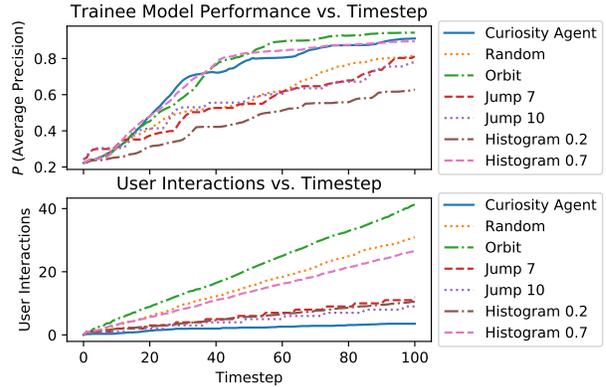}
    \caption{(Top) Plot of AP vs. time-step for 15 simulated episodes. (Bottom) Plot of user interaction vs. time-step for the same episodes. We see that the top strategies perform similarly in AP, but vary vastly in terms of number of user interactions required to get there.}
    \label{fig:sim_timeseries}
\end{figure}

We first measured the performance of the trainee model, as tested on all views in the exploration space, as well as the user interactions (ground truth requests), at each time-step, averaged over the 15 trial environments. These plots are shown in Figure~\ref{fig:sim_timeseries}. We can observe that the \textbf{orbit} strategy produces the highest overall trainee performance at the end of the episode---it will consistently provide 100\% coverage of the exploration space. However, since it requires the user to confirm the presence of the object at each time-step during obstruction, it also requires by far the most user interactions. The \textbf{jump} and \textbf{histogram} strategies (\ie{} subsampling) require less user interactions than orbit, and can achieve similar performance.

The curiosity agent, by intelligently deciding when to request ground truth and when to move, produced similar performance with more than 10 times fewer required user interactions as orbit, and 3-4 times less than \textbf{jump}. This corresponds to the qualitative behaviors of the curiosity agent---we have observed that the learned agent will continue to train on a particular view until most false positives and false negatives are removed, choosing to spend its limited time-steps on particularly problematic views. Furthermore, it is less likely to request user interaction when the view is obstructed.

\begin{figure}[!htb] \centering
    \includegraphics[width=1\linewidth]{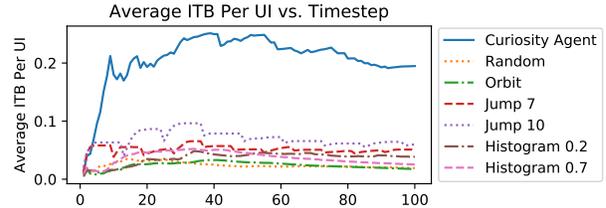}
    \caption{Plot of $\overline{ITB}_{u_{(1,i)}}$ per time-step $i$. We see that the Curiosity Agent produces a much higher value per interaction throughout the sequence.}
    \label{fig:sim_avg_itb}
\end{figure}

We can see the benefit of this behavior more clearly by observing $\overline{ITB}$, rather than raw AP or user interactions. Using Equation~\ref{eqn:meanitb}, we computed the $\overline{ITB}_{u_{(1,i)}}$ for time-steps $i=[1,100]$. This plot is shown in Figure~\ref{fig:sim_avg_itb}.

We observe that when it comes to $\overline{ITB}$, the curiosity agent dominates the baseline strategies by 3.27 times the next best strategy, \textbf{jump 10}. Table~\ref{tab:simulation_results} summarizes these results quantitatively. Versus \textbf{random}, the curiosity agent produces more than 10 times $\overline{ITB}$ per user interaction. While using the curiosity agent does not necessarily produce the best performing model, \textit{it requires the least amount of user input} to get there.

\begin{table}[h]
\resizebox{\linewidth}{!}{%
  \begin{tabular}{ c | c | c | c | c }
    \hline
    \textbf{Strategy} & $P_{100}$& $\sum_{x=0}^{100} u_x $ &  $\overline{ITB}_{u_{(1,100)}}$ & Imp. vs Rand. \\ \hline
    Curiosity Agent & 0.911 & \textbf{3.53} & \textbf{0.195} &  \textbf{10.2x} \\ \hline
    Random & 0.814 & 30.9 & 0.0191 & 1.0x \\ \hline
    Orbit & \textbf{0.943} & 41.3 & 0.0175 &  0.916x   \\ \hline
    Jump 7 & 0.812 & 12.0 & 0.0475 & 2.49x   \\ \hline
    Jump 10 & 0.778 & 9.0 & 0.0597 & 3.12x  \\ \hline
    Histogram 0.2 & 0.628 & 10.5 & 0.0388 & 2.03x  \\ \hline
    Histogram 0.7 & 0.896 & 26.6 & 0.0253 & 1.32x  \\ \hline
  \end{tabular}}
\caption{Comparison of final model performance, number of user interactions, mean incremental training benefit, and $\overline{ITB}$ relative to \textbf{random} strategy. We see that a trained curiosity agent is more than 10 times more effective in utilizing user interactions than \textbf{random}, and more than 3 times compared to the next best baseline.}
   \label{tab:simulation_results}
\end{table}

\vspace{-0.5em}

\section{Results - Field Experiment}
\label{sec:experiments_field}
We validated our curiosity agent on real-world imagery. In doing so, we both test our curiosity agent's generalizability, and its strategy's suitability to real-world subjects. Note that the curiosity agent here is identical to the one in Section~\ref{sec:experiments_simulation}, and \textit{was trained purely on simulated scenes}.

\begin{figure}[!htb] \centering
    \includegraphics[width=0.9\linewidth]{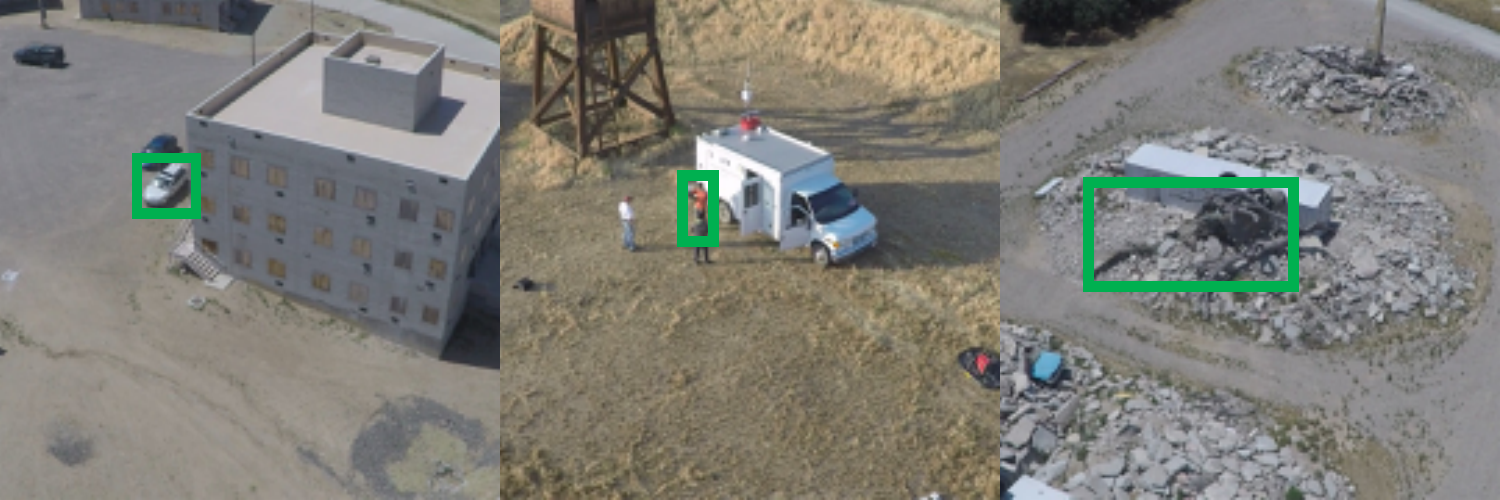}
    \caption{Sample views from three field scenarios with three different subjects (car, person, helicopter wreckage). All three scenarios exhibit views with varying degrees of obstruction and background clutter, with the first two explicitly created to have total obstruction and objects of the same class within the views. The helicopter wreckage represents a scene where the shape of the object varies at different view angles.}
    \label{fig:screenshots}
\end{figure}

We recorded several scenes (Figure~\ref{fig:screenshots}) with similar parameters to the simulated ones described in Section~\ref{sec:problem}. We again set $\Delta \alpha = 12\degree$, and $\theta = 30\degree$ to match the viewpoint of our training scenes. For the car and helicopter scenes, $d = 60$m. We reduced $d$ to 20m for the person scene, as the subject was only several pixels in width at 60 meters.

\begin{figure}[!htb] \centering
    \includegraphics[width=1\linewidth]{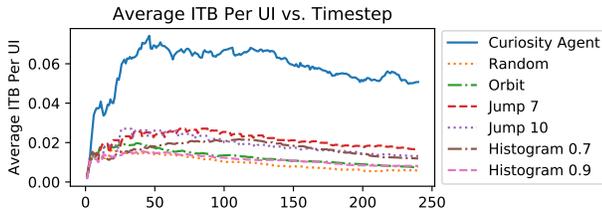}
    \caption{Plot of average $\overline{ITB}_{u_{(1,i)}}$ per $i$. We see that the Curiosity Agent produces a higher value per interaction, just as in simulation.}
    \label{fig:sim_avg_itb_field}
\end{figure}

By flying the exploration space and capturing each view using a drone, we were able to re-create arbitrary episodes with different strategies. We set the time budget $T=240$---at 2.5 seconds per training or movement step, this approximates the 10 minute flight time of our Jetson TX2-equipped drone.  To remove the effect of start position, we evaluated each of the 3 scenes in 5 separate episodes, starting at evenly spaced views around the orbit. We measured the resulting trainee performance and user interactions, and averaged them over the 15 total episodes. Figure~\ref{fig:sim_avg_itb_field} shows the user interaction effectiveness vs. time-step for each of the strategies, and Table~\ref{tab:simulation_results_field} quantifies these results.


\begin{table}[!htb]
\resizebox{\linewidth}{!}{%
  \begin{tabular}{ c | c | c | c | c }
    \hline
    \textbf{Strategy} & $P_{240}$& $\sum_{x=0}^{240} u_x $ &  $\overline{ITB}_{u_{(1,240)}}$ & Imp vs Rand. \\ \hline
    Curiosity Agent & 0.520 & \textbf{8.07} & \textbf{0.0507} &  \textbf{8.82x} \\ \hline
    Random & 0.506 & 68.7 & 0.00575 & 1.0x \\ \hline
    Orbit & \textbf{0.797} & 89.6& 0.00767 &  1.33x   \\ \hline
    Jump 7 & 0.613 & 30.0 & 0.0167 & 2.90x   \\ \hline
    Jump 10 & 0.404  & 22.0 & 0.0131 & 2.28x  \\ \hline
    Histogram 0.7 & 0.319 & 17.5 & 0.0119 & 2.07x  \\ \hline
    Histogram 0.9 & 0.505 & 48.5 & 0.00814 & 1.42x  \\
  \end{tabular}}
\caption{Comparison of final model performance, number of user interactions, mean incremental training benefit, and $\overline{ITB}$ relative to \textbf{random} strategy for field scenarios. We see that using a trained curiosity agent is more than 8 times more effective in utilizing user interactions than \textbf{random}, and more than 3 times more effective than the next best baseline.}
   \label{tab:simulation_results_field}
\end{table}

We note that \textbf{orbit} dominates when it comes to $P$, more so than in the simulated cases. This is especially true for the helicopter scene, where the subject is vastly different in appearance depending on view, and a brute-force annotation strategy is optimal. We believe decreasing $w_p$ in the reward function would help close this gap for such scenes, at the cost of more user interactions. Nevertheless, we observe that when it comes to $\overline{ITB}$, the curiosity agent again dominates the baseline strategies, by 3.04 times the next best strategy, \textbf{jump 7}. Versus \textbf{random}, the curiosity agent produces more than 8 times $\overline{ITB}$ per user interaction.



\section{Potential Extensions}
\label{sec:extensions}

\textbf{Multiple Classes}: We limited the object detector to single-class, as we believed this would be sufficient to prove the concept of autonomous curiosity. The SSD model in the trainee is capable of multi-class detection, provided that the appropriate annotations are provided, and multiple tracker instances are instantiated per object. We do have to represent these classes in the agent state. For a small number of classes (\textasciitilde 1-5), we can append additional channels of confidence-weighted BBoxes (Figure~\ref{fig:staterep}), one for each class. For more classes, we can represent each class as a binary encoding of such channels, so that the number of classes that can be represented in the state is $2^N$, where $N$ is the number of BBox channels, using value (0-255) for confidence.

\textbf{Increased Range-of-Motion}: We can extend to more realistic drone motions by expanding the number of actions available to the agent, and using continuous control rather than discrete steps. The latter can be done using policy gradient methods~\cite{Hwangbo2017ControlLearning,Koch2018ReinforcementControl}, but will require care in how often reward is computed to maintain realistic training times. Both will break the assumption that each step has the same time cost, and require time to be a part of the reward function.

\section{Conclusion}
\label{sec:conclusion}

In this paper, we examined how a robotic agent with movement capability, such as a drone, can assist in human-in-the-loop training. In effect, can we train an agent to be curious and seek out new training data on its own, requesting human assistance only when necessary? We proposed a deep reinforcement learning approach that performs both the tasks of data selection (active learning) and data search through movement (active perception). In our experiments, a trained curiosity agent shows consistently better effectiveness (more than 3x) in using human interactions to train a trainee object detector than untrained baselines. This was shown on both simulated and real scenes. We believe this approach is a step forwards in enabling robotic agents to be collaborative, rather than passive, students of their human operators, reducing overall human mental load. 

{\small
\bibliographystyle{ieee}
\bibliography{Mendeley}

\begin{thebibliography}{10}\itemsep=-1pt

\bibitem{Bayoumi2015EfficientLearning}
A.~Bayoumi and M.~Bennewitz.
\newblock {Efficient Human Following Using Reinforcement Learning}.
\newblock {\em Workshop on Machine Learning in Planning and Control of Robot
  Motion}, 2015.

\bibitem{Bearman2016WhatsSupervision}
A.~Bearman, O.~Russakovsky, V.~Ferrari, and L.~Fei-Fei.
\newblock {What's the Point: Semantic Segmentation with Point Supervision}.
\newblock In {\em ECCV}, 2016.

\bibitem{Bengio2009CurriculumLearning}
Y.~Bengio, U.~J{\'{e}}r{\^{o}}me~Louradour, R.~Collobert, and J.~Weston.
\newblock {Curriculum Learning}.
\newblock In {\em ICML}, 2009.

\bibitem{Caicedo2015}
J.~C. Caicedo and S.~Lazebnik.
\newblock {Active Object Localization with Deep Reinforcement Learning}.
\newblock In {\em ICCV}, 2015.

\bibitem{Fang2017}
M.~Fang, Y.~Li, and T.~Cohn.
\newblock {Learning how to Active Learn: A Deep Reinforcement Learning
  Approach}.
\newblock In {\em Proceedings of the 2017 Conference on Empirical Methods in
  Natural Language Processing (EMNLP)}, pages 606--616, 2017.

\bibitem{Guisti2016}
A.~Guisti, J.~Guzzi, D.~C. Cire{\c{s}}an, H.~Fang-Lin, J.~Rodriguez,
  F.~Fontana, M.~Faessler, C.~Forster, J.~Schmidhuber, G.~Di~Caro,
  D.~Scaramuzza, and L.~M. Gambardella.
\newblock {A Machine Learning Approach to Visual Perception of Forest Trails
  for Mobile Robots}.
\newblock {\em Robotics and Automation Letters}, 1(2):661--667, 2016.

\bibitem{Hausknecht2015DeepMDPs}
M.~Hausknecht and P.~Stone.
\newblock {Deep Recurrent Q-Learning for Partially Observable MDPs}.
\newblock {\em ArXiv e-prints}, abs/1507.0, 2015.

\bibitem{Hwangbo2017ControlLearning}
J.~Hwangbo, I.~Sa, R.~Siegwart, and M.~Hutter.
\newblock {Control of a Quadrotor with Reinforcement Learning}.
\newblock {\em IEEE Robotics and Automation Letters}, 2(4):1--8, 2017.

\bibitem{Jain2016ClickClicks}
S.~D. Jain and K.~Grauman.
\newblock {Click Carving: Segmenting Objects in Video with Point Clicks}.
\newblock In {\em HCOMP}, 2016.

\bibitem{Jayaraman2016Look-aheadMotion}
D.~Jayaraman and K.~Grauman.
\newblock {Look-ahead before you leap: End-to-end active recognition by
  forecasting the effect of motion}.
\newblock In {\em ECCV}, 2016.

\bibitem{Jayaraman2018}
D.~Jayaraman and K.~Grauman.
\newblock {Learning to Look Around: Intelligently Exploring Unseen Environments
  for Unknown Tasks}.
\newblock In {\em CVPR}, 2018.

\bibitem{Juliani2018Unity:Agents}
A.~Juliani, V.-P. Berges, E.~Vckay, Y.~Gao, H.~Henry, M.~Mattar, and D.~Lange.
\newblock {Unity: A General Platform for Intelligent Agents}.
\newblock {\em ArXiv e-prints}, abs/1809.0, 2018.

\bibitem{Koch2018ReinforcementControl}
W.~Koch, R.~Mancuso, R.~West, and A.~Bestavros.
\newblock {Reinforcement Learning for UAV Attitude Control}.
\newblock {\em ArXiv e-prints}, abs/1804.0:1--13, 2018.

\bibitem{Lample2016PlayingLearning}
G.~Lample and D.~S. Chaplot.
\newblock {Playing FPS Games with Deep Reinforcement Learning}.
\newblock {\em arXiv e-prints}, abs/1609.0, 2016.

\bibitem{Liu2016}
W.~Liu, D.~Anguelov, D.~Erhan, C.~Szegedy, S.~Reed, C.~Y. Fu, and A.~C. Berg.
\newblock {SSD: Single Shot Multibox Detector}.
\newblock In {\em ECCV}, 2016.

\bibitem{Luo2018End-to-endLearning}
W.~Luo, P.~Sun, F.~Zhong, W.~Liu, T.~Zhang, and Y.~Wang.
\newblock {End-to-end Active Object Tracking and Its Real-world Deployment via
  Reinforcement Learning}.
\newblock In {\em ICML}, 2018.

\bibitem{Malmir2015}
M.~Malmir, K.~Sikka, S.~Diego, S.~Diego, C.~Science, S.~Diego, D.~Forster,
  J.~Movellan, and G.~W. Cottrell.
\newblock {Deep Q-learning for Active Recognition of GERMS: Baseline
  performance on a standardized dataset for active learning}.
\newblock In {\em BMVC}, 2015.

\bibitem{Martinez-Cantin2007ActiveUncertainty}
R.~Martinez-Cantin, N.~d. Freitas, A.~Doucet, and J.~A. Castellanos.
\newblock {Active Policy Learning for Robot Planning and Exploration under
  Uncertainty}.
\newblock In {\em Robotics: Science and Systems}, 2007.

\bibitem{Mathe2016}
S.~Mathe, A.~Pirinen, and C.~Sminchisescu.
\newblock {Reinforcement Learning for Visual Object Detection}.
\newblock In {\em CVPR}, 2016.

\bibitem{Mettes2016SpotProposals}
P.~Mettes, J.~C. Van~Gemert, and C.~G.~M. Snoek.
\newblock {Spot On: Action Localization from Pointly-Supervised Proposals}.
\newblock In {\em ECCV}, 2016.

\bibitem{Mirowski2017LearningEnvironments}
P.~Mirowski, R.~Pascanu, F.~Viola, H.~Soyer, A.~J. Ballard, A.~Banino,
  M.~Denil, R.~Goroshin, L.~Sifre, K.~Kavukcuoglu, D.~Kumaran, and R.~Hadsell.
\newblock {Learning to Navigate in Complex Environments}.
\newblock In {\em ICLR}, 2017.

\bibitem{Papadopoulos2016WeVerification}
D.~P. Papadopoulos, J.~R.~R. Uijlings, F.~Keller, and V.~Ferrari.
\newblock {We Don't Need No Bounding-Boxes: Training Object Class Detectors
  Using Only Human Verification}.
\newblock In {\em CVPR}, 2016.

\bibitem{Papadopoulos2017}
D.~P. Papadopoulos, J.~R.~R. Uijlings, F.~Keller, and V.~Ferrari.
\newblock {Training Object Class Detectors with Click Supervision}.
\newblock In {\em CVPR}, 2017.

\bibitem{Pathak2017Curiosity-drivenPrediction}
D.~Pathak, P.~Agrawal, A.~A. Efros, and T.~Darrell.
\newblock {Curiosity-driven Exploration by Self-supervised Prediction}.
\newblock In {\em ICML}, Sydney, 2017.

\bibitem{Pestana2014ComputerVehicles}
J.~Pestana, J.~L. Sanchez-Lopez, S.~Saripalli, and P.~Campoy.
\newblock {Computer Vision Based General Object Following For GPS-Denied
  Multimotor Unmanned Vehicles}.
\newblock In {\em IEEE American Control Conference}, 2014.

\bibitem{Ranzato2014}
M.~Ranzato.
\newblock {On Learning Where To Look}.
\newblock {\em ArXiv e-prints}, abs/1405.5, 2014.

\bibitem{Redmon2015}
J.~Redmon, S.~Divvala, R.~Girshick, and A.~Farhadi.
\newblock {You Only Look Once: Unified, Real-Time Object Detection}.
\newblock {\em arXiv:1506.02640}, 2015.

\bibitem{Sadeghi2016CAD2RL:Image}
F.~Sadeghi and S.~Levine.
\newblock {CAD2RL: Real Single-Image Flight without a Single Real Image}.
\newblock In {\em Robotics: Science and Systems Conferece}, 11 2016.

\bibitem{Sener2017}
O.~Sener and S.~Savarese.
\newblock {A Geometric Approach to Active Learning for Convolutional Neural
  Networks}.
\newblock In {\em ICLR}, 2018.

\bibitem{Settles2010}
B.~Settles.
\newblock {Active Learning Literature Survey}.
\newblock Technical report, University of Wisconsin-Madison, 2010.

\bibitem{Tai2016MobileLearning}
L.~Tai and M.~Liu.
\newblock {Mobile Robots Exploration Through CNN-based Reinforcement Learning}.
\newblock {\em Robotics and Biomimetics}, 2016.

\bibitem{Teng2018ClickBAIT:Networks}
E.~Teng, J.~D. Falc{\~{a}}o, R.~Huang, and B.~Iannucci.
\newblock {ClickBAIT: Click-based Accelerated Incremental Training of
  Convolutional Neural Networks}.
\newblock In {\em AIPR (forthcoming)}, 2018.

\bibitem{Valmadre2017}
J.~Valmadre, L.~Bertinetto, J.~Henriques, A.~Vedaldi, and P.~H.~S. Torr.
\newblock {End-to-end Representation Learning for Correlation Filter Based
  Tracking}.
\newblock In {\em CVPR}, 2017.

\bibitem{vanHasselt2016DeepQ-learning}
H.~van Hasselt, A.~Guez, and D.~Silver.
\newblock {Deep Reinforcement Learning with Double Q-learning}.
\newblock In {\em AAAI}, 9 2016.

\bibitem{Wang2017}
K.~Wang, D.~Zhang, Y.~Li, R.~Zhang, and L.~Lin.
\newblock {Cost-Effective Active Learning for Deep Image Classification}.
\newblock {\em IEEE Transactions on Circuits and Systems for Video Technology},
  pages 1--10, 2017.

\bibitem{Wang2016DuelingLearning}
Z.~Wang, T.~Schaul, M.~Hessel, M.~Com, H.~Van~Hasselt, and M.~Lanctot.
\newblock {Dueling Network Architectures for Deep Reinforcement Learning}.
\newblock In {\em ICML}, 2016.

\bibitem{Xu2016}
N.~Xu, B.~Price, S.~Cohen, J.~Yang, and T.~Huang.
\newblock {Deep Interactive Object Selection}.
\newblock In {\em CVPR}, 2016.

\bibitem{Yao2012InteractiveDetection}
A.~Yao, J.~Gall, C.~Leistner, and L.~Van~Gool.
\newblock {Interactive Object Detection}.
\newblock In {\em CVPR}, 2012.

\bibitem{Yeung2016End-to-endVideos}
S.~Yeung, O.~Russakovsky, G.~Mori, and L.~Fei-Fei.
\newblock {End-to-end Learning of Action Detection from Frame Glimpses in
  Videos}.
\newblock In {\em CVPR}, 2016.

\end{thebibliography}
}
\end{document}